\documentclass[10pt]{article}

\usepackage[margin=1in]{geometry}
\usepackage{graphicx}
\usepackage{microtype}
\usepackage[hidelinks]{hyperref}
\usepackage{titlesec}
\usepackage{caption}
\usepackage{float}

\titleformat{\section}{\large\bfseries}{\thesection}{0.6em}{}
\titlespacing*{\section}{0pt}{0.75em}{0.30em}
\title{Comment on arXiv:2607.01233:\\ Survivorship Bias in Published-Paper Baselines for Research-Idea Distributions}
\author{Fredrik A. Dahl}
\date{}

\begin{document}
\maketitle
\vspace{-1.5em}

\begin{abstract}
Chen, Zhao, and Cohan introduce a valuable distributional evaluation of LLM-generated research ideas. This comment raises a narrower identification concern: their human baseline consists of published papers, whereas the LLM baseline consists of one-shot proposals. If bridge-like or synthesis-like ideas are relatively easy to generate but relatively unlikely to survive publication, then the published human baseline will understate their prevalence in the unseen human idea pool. The observed human--LLM gap may therefore be partly, or even largely, a consequence of survivorship bias.
\end{abstract}

\section{Survivorship bias, not merely refinement}

The main empirical result in \cite{chen2026} is striking: LLM-generated ideas are far more concentrated in bridge-like opportunities and synthesis/unification methods than the published human-paper reference distribution. I do not dispute the descriptive finding. The concern is what the human reference distribution represents.

A published paper is not a random draw from human first-pass ideation. It is a survivor. Many possible projects are mentioned in lab meetings, sketched in notes, tried briefly, rejected by collaborators, abandoned after preliminary failures, declined by funders, or screened out by reviewers. The human statistic in the paper is therefore a distribution conditional on survival into the published literature. That is a classic survivorship-bias setting.

This survivorship bias is especially relevant because it is not plausibly neutral across the paper's taxonomy. Bridge-like ideas are easy to formulate: connect two adjacent literatures, apply one method to another setting, or synthesize nearby strands of work. 
Weak versions of such ideas are also easy to reject. They may be obvious, boilerplate, incremental, or insufficiently differentiated from prior work. 
Indeed, their limited scientific value is precisely why the article frames there abundance among LLM-generated ideas as problematic.
Thus, if bridge and boilerplate ideas have low publication probability, they should be rare among published human papers even if they are common in human zero-shot brainstorming.

This point does not require assuming that ideas change taxonomy during development. Even if every published paper is labeled perfectly, and even if its extracted idea faithfully represents the final contribution, survivorship bias remains: categories with low survival probability are underrepresented among survivors.

\section{A counterexample distribution}

\begin{figure}[H]
\centering
\includegraphics[width=0.72\linewidth]{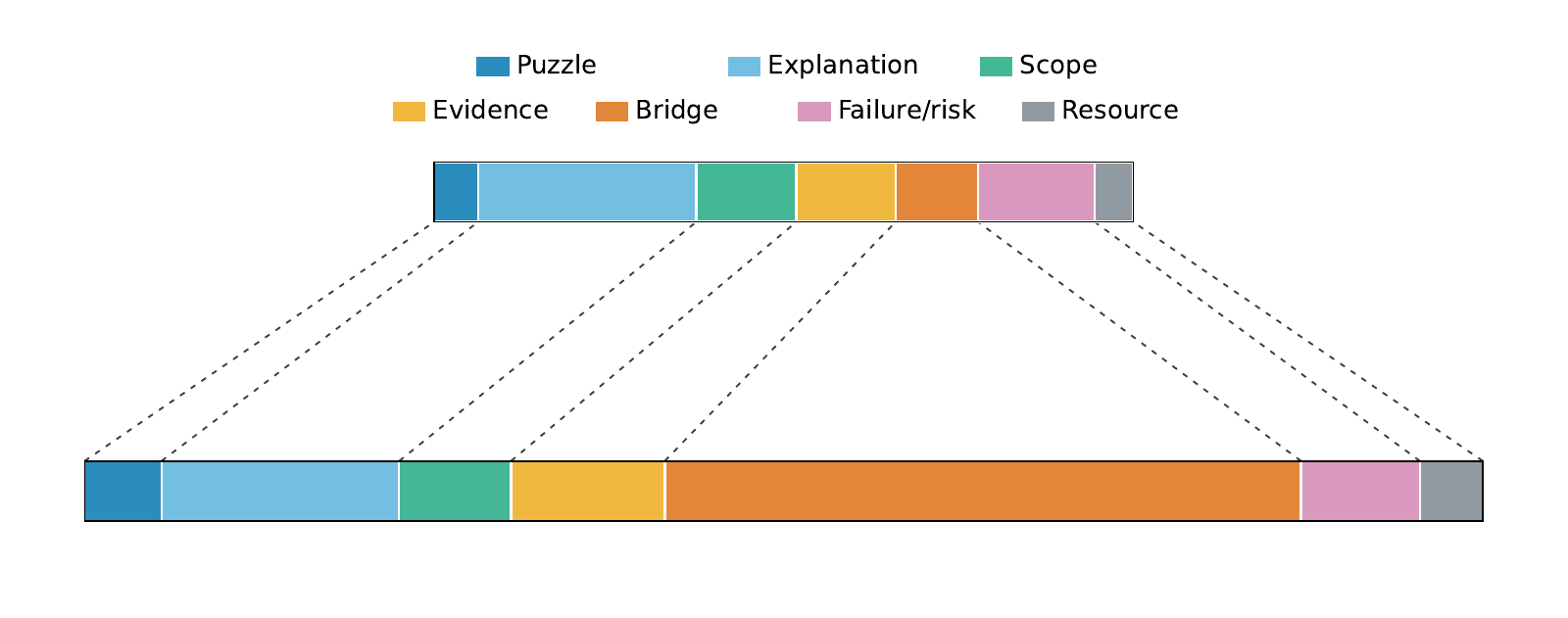}
\caption{Schematic survivorship-bias alternative. The lower row is a hypothesized zero-shot human idea distribution, similar to the observed LLM distribution. The upper row is a hypothetical observed surviving distribution after publication selection, similar to the observed human distribution. Dotted lines connect corresponding taxonomy categories.}
\label{fig:survivorship}
\end{figure}

Figure~\ref{fig:survivorship} illustrates the survivorship-bias alternative. The lower row is a hypothesized zero-shot human idea distribution, drawn to be similar to the bridge-heavy LLM distribution observed in \cite{chen2026}. The dotted lines indicate differential survival into the published record. The upper row is the resulting hypothetical survivor distribution; it is bridge-light and similar to the observed human-paper distribution. The figure is schematic, not an estimate.

The key point is that this counterexample is not exotic. It follows from the paper's own interpretation of bridge-heavy output as a relatively generic and narrow mode of ideation. If that diagnosis is right, the same property should make many bridge-heavy ideas less publishable. Their scarcity in the human publication statistic is then not surprising; it is what survivorship bias predicts.

In the terminology of inverse-probability weighting, ideas with low publication probability would receive large correction factors when reconstructing the pre-publication idea distribution from the published-paper distribution. If bridge or boilerplate ideas have low publication probability, their inverse-publication weights would be large. The figure does not represent a numerical estimate, since the required survival probabilities are unobserved. It is simply the standard selection-bias logic showing why the published-paper distribution should not be treated as the unfiltered human idea distribution.

\section{What comparison is needed}

The clean comparison is stage-matched. Human researchers should be given the same reconstructed prior-work packets and asked for first-pass ideas before any development or publication filter. If those first-pass human ideas are also bridge-heavy, survivorship bias would explain much of the reported gap. If they are not, the claim of an LLM-specific research-taste gap would be stronger. 
Other useful checks would sample the human ``idea graveyard''; rejected submissions, abandoned drafts, unfunded proposals, etc, to estimate the pre-survival idea distribution.

Until such evidence is available, the safest conclusion is narrower than the strong research-taste interpretation. The paper shows that one-shot LLM proposals differ from published human papers. It does not, by itself, rule out the possibility that the difference is a survivorship-bias artifact created by comparing pre-selection LLM ideas with post-selection human papers.

The present commentary, btw, merely bridges statistics methods with the article in question in a boilerplate way. Although it was written with the support of an LLM, the idea was human.

\end{document}